\documentclass{article}
\usepackage{ijcai22}
\usepackage{times}
\usepackage{epsfig}
\usepackage{graphicx}
\usepackage{subfigure}  
\usepackage{amsmath}
\usepackage{amssymb}

\usepackage{color}
\definecolor{citecolor}{RGB}{119,185,0} 

\usepackage{soul}
\usepackage{url}
\usepackage[utf8]{inputenc}
\usepackage[small]{caption}
\usepackage{multirow}
\usepackage{multicol}
\usepackage{comment}
\usepackage{booktabs}
\usepackage[title]{appendix}

\usepackage{xcolor}

\usepackage{pifont}

\def\ie{\emph{i.e.}}

\newlength\savewidth

\begin{document}

\title{Learning Prototype via Placeholder for Zero-shot Recognition}

\author{
Zaiquan Yang$^1$\and
Yang Liu$^1$\and
Wenjia Xu$^2$\and 
Chong Huang$^1$\and 
Lei Zhou$^1$\and 
Chao Tong$^1$\thanks{Corresponding author}
\affiliations
$^1$Beihang University\\
$^2$Beijing University of Posts and Telecommunications\\
\emails
zaiquanyangcat@gmail.com,
\{wickerboy, chonghuang, leizhou, tongchao\}@buaa.edu.cn, \\
xuwenjia16@mails.ucas.ac.cn
}


\maketitle

\begin{abstract}

Zero-shot learning (ZSL) aims to recognize unseen classes by exploiting semantic descriptions shared between seen classes and unseen classes. 
Current methods show that it is effective to learn visual-semantic alignment by projecting semantic embeddings into the visual space as class prototypes. However, such a projection function is only concerned with seen classes. When applied to unseen classes, the prototypes often perform suboptimally due to domain shift. 
In this paper, we propose to learn prototypes via placeholders, termed LPL, to eliminate the domain shift between seen and unseen classes. Specifically, we combine seen classes to hallucinate new classes which play as placeholders of the unseen classes in the visual and semantic space. Placed between seen classes, the placeholders encourage prototypes of seen classes to be highly dispersed. And more space is spared for the insertion of well-separated unseen ones. Empirically, well-separated prototypes help counteract visual-semantic misalignment caused by domain shift.
Furthermore, we exploit a novel semantic-oriented fine-tuning to guarantee the semantic reliability of placeholders. 
Extensive experiments on five benchmark datasets demonstrate the significant performance gain of LPL
over the state-of-the-art methods. 
Code is available at https://github.com/zaiquanyang/LPL.

\end{abstract}

\section{Introduction}
Inspired by the human cognitive system, zero-shot learning (ZSL) was proposed to identify unseen classes
by utilizing semantic embeddings (e.g., attributes
\cite{lampert2009learning} 
or text descriptions
\cite{reed2016learning}
) to transfer knowledge from seen domain to unseen domain. The ZSL can be categorized into conventional and generalized settings according to the different classes that a model sees in the test phase. In conventional ZSL, the test images from the unseen domain will be recognized. And the more challenging generalized ZSL (GZSL)~\cite{xian2017zero} aims to predict the test images belonging to both the seen and unseen domains.

\begin{figure}[t]
    \centering
    \includegraphics[width=0.8\columnwidth]{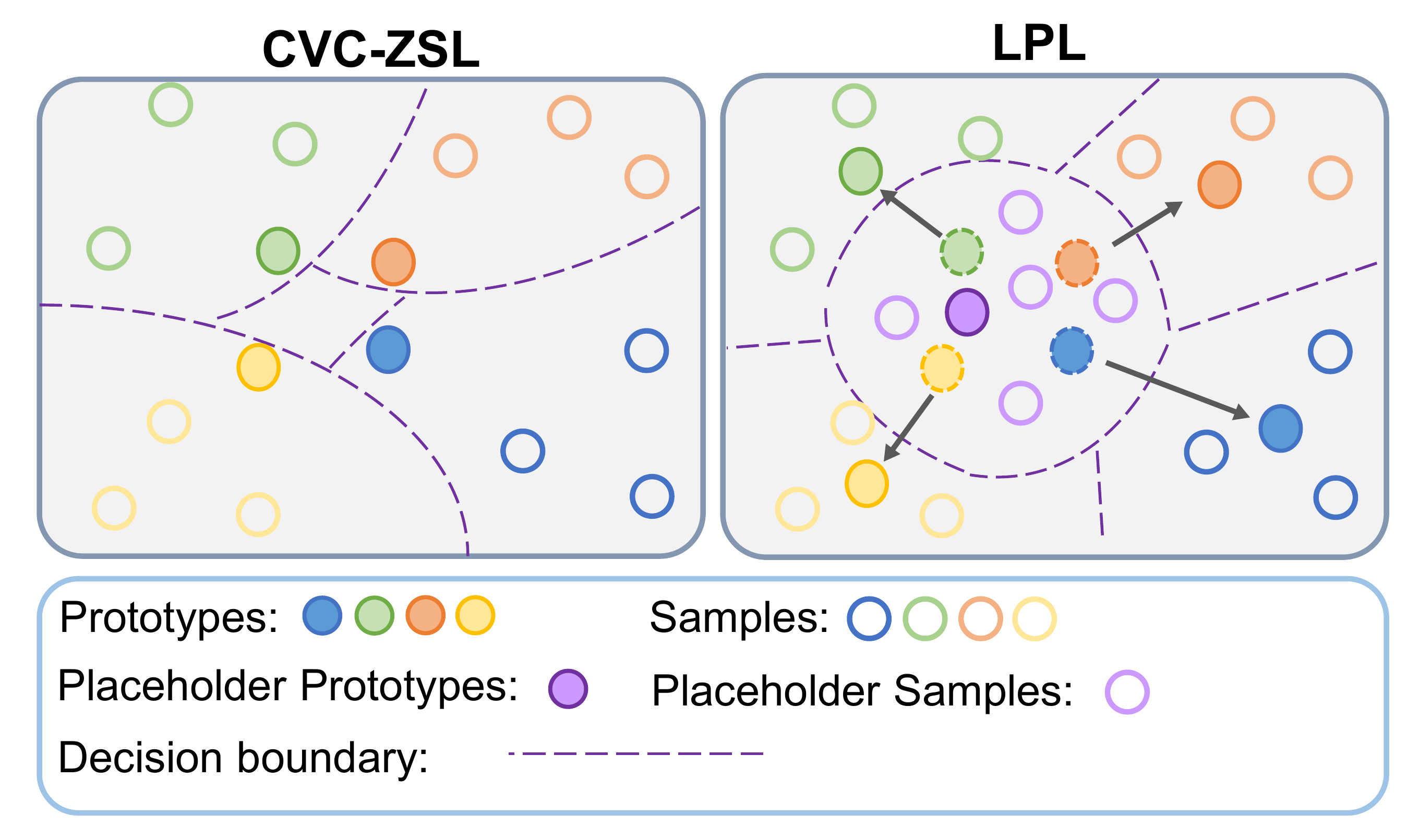}
     \caption{In CVC-ZSL, seen classes prototypes learned by mapping seen classes semantic embeddings into visual space are generally located in confined space as shown in the left image. By learning placeholders (purple circles) placed between seen classes (circles marked with different colors), LPL can obtain well-separated seen classes prototypes and spares more space for unseen ones to counter domain shift as shown in the right image. 
      }
\label{fig:motivation}
\end{figure}

Domain shift~\cite{fu2015transductive} is an intractable problem in ZSL, due to the underlying difference between data distributions of seen domain and unseen domain. Only built on seen domain, the visual-semantic alignment is often distorted and loses desired discrimination especially for semantically similar categories of unseen domain. 
Early ZSL methods~\cite{bucher2016improving} learn projecting visual features into semantic space to improve semantic representative capability of the features and eliminate the domain shift. However, some pioneers argue that taking semantic space as the projection space is less discriminating due to the hubness problem~\cite{radovanovic2010hubs}. When high-dimensional visual features are mapped to a low-dimensional semantic space, the shrink of feature space would aggravate the hubness problem that some instances in the high-dimensional space become the nearest neighbors of a large number of instances~\cite{liu2020information}.
To tackle these problems, CVC-ZSL\cite{li2019rethinking} proposes mapping semantic embeddings to visual space and treats the projected results as class prototypes. Though the prototypes of seen classes are highly discriminative, it is still suboptimal due to the lack of unseen classes in the training phase.



Since the deep neural network (DNN) tends to predict with a subset of the most predictive features 
\cite{huang2020self}, 
embedding-based methods 
are limited to learning visual-semantic alignment of seen classes 
and generally obtain prototypes of seen classes located in a confined space as shown in Figure \ref{fig:motivation}~(left). 
Being constricted in such a limited space, prototypes of unseen classes can  easily lose the discrimination due to the disruption of the domain shift and cannot be well adapted to recognize unseen classes in ZSL setting or distinguish the unseen from seen classes in GZSL setting.  
As a result, to cope with the domain shift, the prototypes of unseen classes require more space for highly separable arrangements.

In this work, we propose to learn prototypes via placeholders for zero-shot recognition (\textbf{LPL}) to mitigate the domain shift. Figure \ref{fig:motivation} indicates the motivation of our method. 
Building upon the idea that the unseen class usually share semantics with several seen classes, e.g., zebra is black and white animals~(as gaint panda) with four legs~(as tiger) and horse shape body~(as horse). 
LPL utilizes the combination of seen classes to hallucinate both visual and semantic embeddings for new classes scattering among seen classes.
As shown in Figure \ref{fig:motivation}~(right), taking hallucinated classes as placeholders for unseen classes, LPL learns highly dispersed prototypes of seen classes. Thus, more space is spared to insert prototypes of unseen classes and tackle the impact of domain shift.
Specifically, an effective two-steps hallucination strategy is proposed. First, we blend visual and semantic embeddings of multiple seen classes on a similarity graph respectively and control the hallucination classes to distribute around seen classes without significantly deviating from the original data. Second, to obtain abundant classes as placeholders for unseen classes, we further interpolate between elementary hallucinated classes obtained in the previous step and the original seen classes. To prevent the semantic ambiguity of the hallucinated classes from weakening the effect of placeholders, a semantic-oriented fine-tuning strategy is proposed for preliminary visual-semantic alignment which promotes the feasibility of placeholders.


\textbf{Our contribution} is three-fold: 
\textbf{(1)} 
We propose placeholder-based prototype learning for zero-shot recognition (\textbf{LPL}), which hallucinates new classes playing as placeholders of unseen classes and encourages learning well-separated prototypes for unseen classes recognition.
\textbf{(2)} 
With the proposed semantic-oriented fine-tuning, we prevent generating substandard hallucinated classes that trigger semantic ambiguity, which promotes the faithfulness of the placeholders.
\textbf{(3)} With extensive experiments and ablation study on five benchmarks, we demonstrate that the proposed LPL achieves new state-of-the-art ZSL performance.


\section{Related Works}

\subsection{Visual-Semantic Gap}

Zero-shot learning (ZSL) transfers knowledge from seen classes to unseen by class semantic embeddings. Visual and semantic are two kinds of modal embeddings located on different manifold structures. Thus, there is typically a gap between visual and semantic domains. Thus the crucial task of ZSL is to learn a visual-semantic alignment. Embedding-based methods \cite{xian2017zero,li2019rethinking} learn a common space to bridge the gap. Specifically, CVC-ZSL\cite{li2019rethinking} thinks visual space has highly precious discriminative power and proposes classifying visual features based on prototypes projected from semantic embeddings. Our work also learns classes prototypes by semantic$\rightarrow$visual mapping for classification.

\subsection{Projection Domain Shift}

The problem of domain shift in ZSL is proposed by \cite{fu2015transductive} and known as the projection domain shift. 
The same attributes may have very different visual appearances in terms of seen and unseen classes of ZSL. Thus, the visual-semantic alignment, \ie, the projection function learned from the seen classes, is often distorted when directly applied to the unseen classes. 
SAE \cite{kodirov2017semantic} takes the encoder-decoder paradigm to enforces the reconstruction constraint on seen classes. However, it is less discriminating to project visual features into semantic space.  As the baseline of our work, CVC-ZSL\cite{li2019rethinking} projects semantic embeddings to visual space and treats the projected results as classes prototypes. Due to the lack of unseen classes, the domain shift still cannot be well managed.
Differently, LPL alleviates the domain shift by improving the separability of class-level prototypes. Dispersed prototypes are learned by reserving placeholders for the unseen classes. Placeholder is implemented by class hallucination which has been explored in \cite{zhang2021hallucination} to mitigate the lack of samples for few-shot detection scenario. In this work, with different motivation and implementation, we leverage class hallucination to play a bridge between seen and unseen classes.


\begin{figure*}[t] 
\centering %

\includegraphics[width=0.96\textwidth]{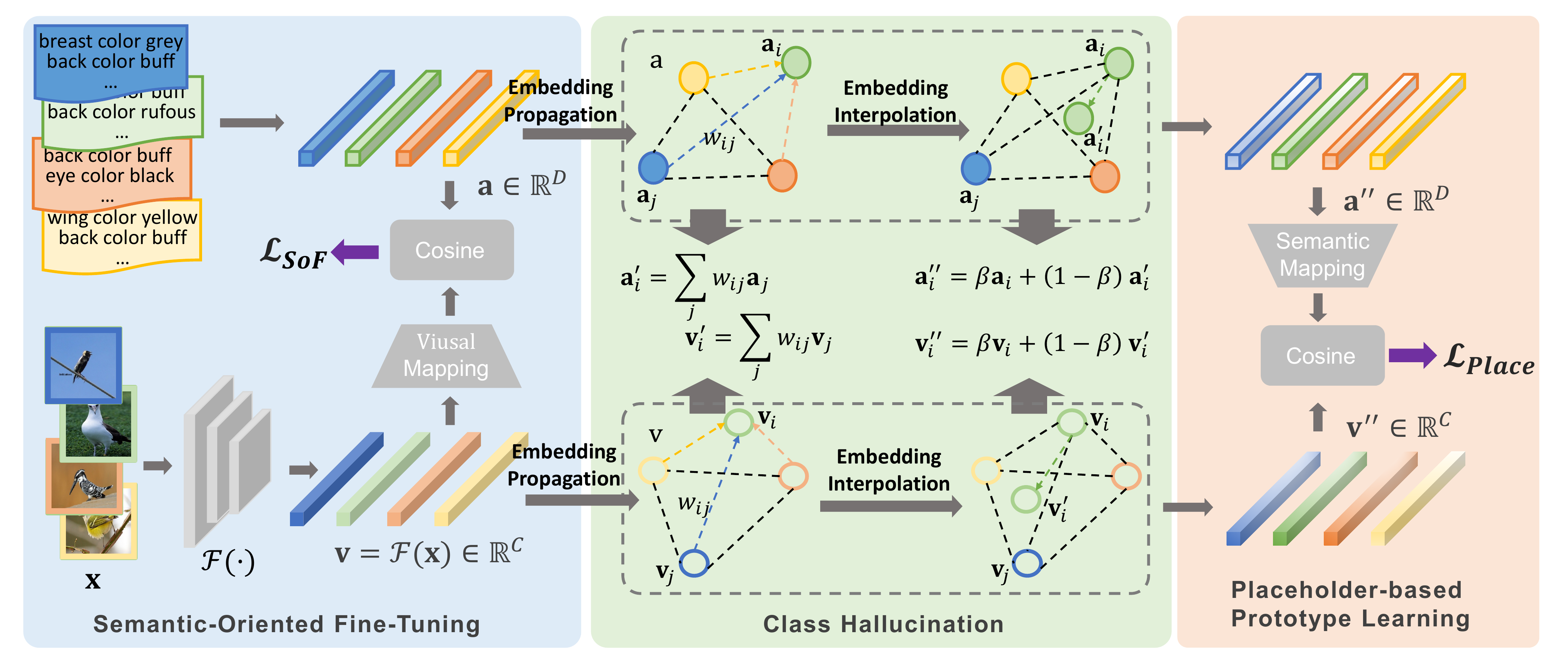}
\caption{ Overview of the proposed framework. Different from learning a plain semantic mapping, LPL learns classes prototypes via hallucinated classes as placeholders of the unseen domain. Also, to prevent substandard hallucination, a preliminary visual-semantic alignment is achieved by semantic-oriented fine-tuning.
} 
\label{Fig.model} 
\end{figure*}

\section{Method}

\subsection{Preliminaries}
We denote the seen dataset as $\mathcal{S} = \{x_{i}, y_{i}\}_{i=1}^{|\mathcal{S}|}$, where $x_{i}$ is an image, $y_{i}$ is its class label in seen classes set $\mathcal{Y}^{s}$ and $|\mathcal{S}|$ is the number of seen images.  Let $\mathcal{U} = \{x^{u}_{i}, y^{u}_{i}\}_{i=1}^{|\mathcal{U}|}$ denotes the unseen dataset, in which the class label $y^{u}_{i}$ is in unseen classes set $\mathcal{Y}^{u}$ and $|\mathcal{U}|$ is the number of unseen images. $\mathcal{Y}^{u}$ is disjoint from the seen classes set $\mathcal{Y}^{s}$. Given an image $x_{i}$, through a visual backbone, the extracted visual feature is denoted as $\bf{v_{i}} \in \mathbb{R}^{C}$.
The number of seen classes and unseen classes are denoted as $L_{s}$ and $L_{u}$, respectively. For the common semantic space transferring information between seen and unseen classes, $\mathcal{A}=\{{\bf{a}}_{k}\}_{k=1}^{L_s+L_u}$  is consisted of class-level semantic attribute vectors, where ${\bf{a}}_{k} \in \mathbb{R}^{D}$. In ZSL, only the seen dataset $\mathcal{S}$ is used during training and the task is to predict the labels of images from unseen classes $\mathcal{Y}^{u}$. For a more realistic and challenging scenario, GZSL is to predict images from both seen and unseen classes, \ie, $\mathcal{Y}^{s} \cup \mathcal{Y}^{u}$.

\subsection{Overview}\label{sec:Overview}

The pipeline of LPL is shown in Figure \ref{Fig.model}. LPL includes two learnable parts: semantic-oriented fine-tuning (SoF) and placeholder-based prototype learning (PPL).
SoF includes the CNN backbone and a visual$\rightarrow$semantic mapping network (denoted as visual mapping) which are jointly optimized and aim to provide a preliminary visual-semantic alignment for credible classes hallucination.
After semantic-oriented fine-tuning, the parameters of the backbone are frozen. Based on seen classes, PPL obtains hallucinated classes that play as placeholders of unseen classes. Then the placeholders are leveraged to learn well-separated classes prototypes by mapping from semantic embeddings.

\subsection{Placeholder-based Prototype Learning}



Besides seen classes, we also hallucinate new classes as input of the semantic mapping network. Playing as placeholders of unseen classes, the hallucination classes prompt dispersed prototypes of seen classes and spare more space for inserting unseen ones. 
In detail, classes hallucination includes two steps: embedding propagation (EP) and embedding interpolation (EI).  EP respectively blends visual/semantic embeddings of multiple classes by implementing propagation on the graph. EI further interpolates the elementary hallucinated visual/semantic embeddings and original ones for more hallucinations, which support placeholder-based prototype learning. We will introduce each operation in the following.

\noindent \textbf{Embedding Propagation.}
We propose to hallucinate new classes with the visual and corresponding semantic embeddings of seen classes. Thus, embedding propagation is implemented on undirected fully connected graphs $G_v$ and graph $G_a$ corresponding to visual and semantic two spaces respectively as shown in Figure \ref{Fig.model}. The nodes of the two graphs are respectively visual and corresponding semantic embeddings.  The edges represent the propagation weight.  Given a batch of data
$
{ \{{\bf{v}}, {\bf{a}} \}} = 
\{
\{{\bf{v}}_{1}, {\bf{a}}_{1}\}, 
\{{\bf{v}}_{2}, {\bf{a}}_{2}\}, 
\cdots, 
\{{\bf{v}}_{k}, {\bf{a}}_{k}\}
\}$ from different seen classes,
embedding propagation outputs a batch of hallucinated classes 
${ 
{ \{\bf{v}^{\prime}, \bf{a}}^{\prime} \}} 
= 
\{ 
\{{\bf{v}}_{1}^{\prime}, {\bf{a}}_{1}^{\prime} \}, 
\{{\bf{v}}_{2}^{\prime}, {\bf{a}}_{2}^{\prime} \}, 
\cdots, 
\{{\bf{v}}_{k}^{\prime}, {\bf{a}}_{k}^{\prime} \} 
\}$  
where visual/semantic embeddings are formulated as follows:
\begin{equation}
    {\bf{v}}_{i}^{\prime} = \sum_{j \in \mathcal{N}^{\prime}_{i}} w_{ij} {\bf{v}}_{j}, 
    \quad
    {\bf{a}}_{i}^{\prime} = \sum_{j \in \mathcal{N}^{\prime}_{i}} w_{ij} {\bf{a}}_{j},
    \label{eq4}
\end{equation}
where $\mathcal{N}^{\prime}_{i}$ is a subset randomly chosen from all neighbor nodes set $\mathcal{N}_{i}$. The number of chosen classes $n$ is a hyper-parameter. Empirically, combining too many neighbor nodes would cause semantic ambiguity and produce less reliable hallucination classes. Considering that visual and semantic embeddings are located in different manifold structures, it is necessary to synchronize the hallucination in visual and semantic spaces. Thus, we apply the same propagation weight $w_{ij}$ on two graphs.


To determine the propagation weight, for the visual space, we firstly compute the raw distance $d_{ij}^{v} = \delta({\bf{v}}_{i}, {\bf{v}}_{j})$ for each pair of visual embeddings. Here $\delta$ denotes a distance function, e.g, cosine distance. To guarantee the convex combinations, we apply softmax function to adjust the  distance $d_{ij} \in [0, 1]$ as follow:
\begin{equation}
    w_{ij}^{v}= 
    \begin{cases}
    \frac{\text{exp}({d_{ij}^{v}} / {\sigma})}
    {\sum_{l\in \mathcal{N}_{i}  } \text{exp} {({d_{il}^{v}} / {\sigma})}}, 
    & i \neq j 
    \\ 
    0, 
    & i=j
    \end{cases}
    \label{eq3}
\end{equation}
where $\sigma$ is the scaling factor and $\mathcal{N}_{i}$ denotes the set composed of neighbor nodes from the same batch. For semantic embeddings space, there are similar operations and we can get the node distance $w_{ij}^{a}$ in semantic space. 
To keep the embedding propagation in a synchronized manner in visual and semantic spaces, 
we  harmonize the propagation weight in two spaces as the final propagation weight  
$w_{ij}= {\left( w_{ij}^{v}+w_{ij}^{a} \right)}/{2}$.

\noindent \textbf{Embedding Interpolation.}
By embedding propagation we obtain hallucination classes located around the pivot of multiple classes. Considering that the distribution of unseen classes is unknown, we should generate as diverse placeholders as possible.
Based on  elementary hallucination classes 
$
{ 
{ \{\bf{v}^{\prime}, \bf{a}}^{\prime} \}} 
= 
\{ 
\{{\bf{v}}_{1}^{\prime}, {\bf{a}}_{1}^{\prime} \}, 
\{{\bf{v}}_{2}^{\prime}, {\bf{a}}_{2}^{\prime} \}, 
\cdots, 
\{{\bf{v}}_{k}^{\prime}, {\bf{a}}_{k}^{\prime} \} 
\}
$   
in the previous step, it is possible to produce more reliable and diverse new classes as placeholders. 

Inspired by manifold mixup \cite{verma2019manifold}, we propose to interpolate between the elementary hallucination classes data and original seen classes data on the graph as shown in Figure \ref{Fig.model}. 
Given a set of original classes data
$
{ \{{\bf{v}}, {\bf{a}} \}} = 
\{
\{{\bf{v}}_{1}, {\bf{a}}_{1}\}, 
\{{\bf{v}}_{2}, {\bf{a}}_{2}\}, 
\cdots, 
\{{\bf{v}}_{k}, {\bf{a}}_{k}\}
\}$ 
and hallucinated classes data 
$
{ 
{ \{\bf{v}^{\prime}, \bf{a}}^{\prime} \}} 
= 
\{ 
\{{\bf{v}}_{1}^{\prime}, {\bf{a}}_{1}^{\prime} \}, 
\{{\bf{v}}_{2}^{\prime}, {\bf{a}}_{2}^{\prime} \}, 
\cdots, 
\{{\bf{v}}_{k}^{\prime}, {\bf{a}}_{k}^{\prime} \} 
\}
$
, we mix up them with very little computational effort in both visual and semantic spaces as follows:
\begin{equation}
\begin{aligned}
{{\bf{v}}_{i}^{\prime \prime}} &=  \beta {\bf{v}}_{i} + (1-\beta) {\bf{v}}_{i}^{\prime}    \\
{{\bf{a}}_{i}^{\prime \prime}} &=  \beta {\bf{a}}_{i} + (1-\beta) {\bf{a}}_{i}^{\prime},
\end{aligned}
\label{eq5}
\end{equation} 
where $\beta \in [0, 1]$ is sampled from Beta distribution $\text{Beta}(\alpha_1, \alpha_2)$. 
Especially, when $\beta=1$ the final hallucination classes are just the original seen classes without any semantic ambiguity. Instead when $\beta=0$ the output classes are just the previous elementary hallucination classes with some semantic ambiguity. On the one hand, using elementary hallucination classes as a pivot, the abundance of placeholders is substantially improved. On the other hand, the ratio factor $\beta$ of original data adjusts the semantic authenticity of hallucination classes.

\noindent \textbf{Visual Feature Classification.}
Adopting episodic training fashion, for each batch of data composed of $k$ seen classes, we can obtain corresponding $k$ hallucinated classes
${ 
{\{\bf{v}^{\prime \prime}, \bf{a}}^{\prime \prime}\}} 
= 
\{ 
\{{\bf{v}}_{1}^{\prime \prime}, {\bf{a}}_{1}^{\prime \prime} \}, 
\{{\bf{v}}_{2}^{\prime \prime}, {\bf{a}}_{2}^{\prime \prime} \}, 
\cdots, 
\{{\bf{v}}_{k}^{\prime \prime}, {\bf{a}}_{k}^{\prime \prime} \} 
\}$
, which play as placeholders of unseen classes and are used for learning visual feature classification. In fact, according to our hallucination strategy, the hallucinated classes are likely to be original seen classes.
Following CVC-ZSL \cite{li2019rethinking}, we learn class prototypes by a semantic$\rightarrow$visual mapping network denoted as semantic mapping in Figure \ref{Fig.model}. For hallucination classes, the classification loss is defined as:
\begin{equation}
    \mathcal{L}_{Place}
    =
    -\log
    \frac{\exp(\cos(h({\bf{a}}_{i}^{\prime \prime}), {\bf{v}}_{i} ^{\prime \prime}))}
    {\sum_{l=1}^{l=k} \exp(\cos(h({\bf{a}}_{l}^{\prime \prime}), {\bf{v}}_{l} ^{\prime \prime}))},
    \label{eq6}
\end{equation}
where $h(\cdot)$ denotes the semantic$\rightarrow$visual mapping function. Also, we measure cosine distance between projected classes prototypes and visual samples.

\subsection{Semantic-oriented Fine-tuning}
According to our motivation, hallucinated classes play as the placeholders of real unseen classes. 
Due to the visual and semantic embeddings located in two different manifold spaces, a favorable visual-semantic alignment is absent. Especially, the visual representations generally contain lots of semantic-unrelated components and show poor intra-class compactness. As a result, the hallucinated classes easily trigger semantic ambiguity and appear unreliable compared to genuine classes. With substandard hallucination, the desired effect of placeholders is weakened.

To mitigate the above challenges, we propose semantic-oriented fine-tuning (SoF) for preliminary visual-semantic alignment. Especially, SoF takes semantic attributes as supervision and fine-tune the backbone followed by a plain visual$\rightarrow$semantic mapping network denoted as visual mapping in Figure \ref{Fig.model}. 
The mapping network is implemented by a linear full-connected layer. 
Denote the weight matrix ${\bf{W}} \in \mathbb{R}^{D \times C}$, where $D$ and $C$ respectively denote the dimension of semantic and visual space. Given the input image $x_i$, visual features are extracted by backbone $\mathcal{F}(\cdot)$ and then projected to semantic space by the transformation matrix $\bf{W}$ for classification. The loss is formulated as:
\begin{equation}
    \mathcal{L}_{SoF}
    =
    -\log
    \frac{\exp(\cos(\mathcal{F}(x_{i}){{\bf{W}}}, {\bf{a}}_{i})}
    {\sum_{l \in \mathcal{Y}^{S}} \exp(\cos(\mathcal{F}(x_{i}){{\bf{W}}}, {\bf{a}}_{l}))},
\end{equation}
which jointly optimizes the backbone and the visual mapping network.

\subsection{Training and Inference}
The training of our model includes two stages as described in Section \ref{sec:Overview}. 
The backbone is fine-tuned guided by semantic attributes in the first stage and frozen in the second stage. For the second stage, we finish learning prototypes with the benefit of hallucination classes.
It should be noted that the whole training does not use any seen classes visual samples or semantic embeddings. For the inference, we use the learned semantic mapping network to obtain unseen classes prototypes for zero-shot recognition as follow:
\begin{equation}
    \hat{y}=\underset{y \in \mathcal{Y}^{U}}{\arg \max } \cos \left(h({\bf{a}}_{y}), \mathcal{F}(x_{i})\right).
\end{equation}
For GZSL, to mitigate the bias towards seen classes, we apply calibrated stacking (CS) 
to reduce seen class scores by a calibration factor $\delta$ as follow:
\begin{equation}
    \hat{y}=\underset{y \in \mathcal{Y}^{U} \cup \mathcal{Y}^{S}}{\arg \max }\left(\cos \left(h\left({\bf{a}}_{y}\right), \mathcal{F}\left(x_{i}\right)\right)-\delta \mathbb{I}\left[y \in \mathcal{Y}^{S}\right]\right).
\end{equation}

\section{Experiment}


\renewcommand\arraystretch{1.0}
\begin{table*}[ht]
		\centering
		\resizebox{0.98\textwidth}{!}
		{
			\begin{tabular}{l|cccc|cccc|cccc|cccc|cccc}
			\toprule[2pt]
				\multirow{2}*{Method}
				&\multicolumn{4}{c|}{AWA2}
				&\multicolumn{4}{c|}{CUB}     
				&\multicolumn{4}{c|}{SUN}
				&\multicolumn{4}{c|}{FLO}
				&\multicolumn{4}{c}{APY}
				\\
				& $T$ & ${U}$ & ${S}$ & ${H}$ 
				& $T$ & ${U}$ & ${S}$ & ${H}$
				& $T$ & ${U}$ & ${S}$ & ${H}$
				& $T$ & ${U}$ & ${S}$ & ${H}$
				& $T$ & ${U}$ & ${S}$ & ${H}$
				\\
				\midrule
				
		
				f-VAEGAN \textit{CVPR 2019}~\cite{xian2019f}
				&70.3&57.1&76.1&65.2
				&72.9&63.2&75.6&68.9
				&\textbf{\color{blue}65.6} &\textbf{\color{blue}50.1} &37.8&43.1
				&70.4&63.3 &\textbf{\color{red}92.4} &\textbf{\color{blue}75.1}
				&--&--&--&--
				\\
				
				CADA-VAE \textit{CVPR 2019}~\cite{schonfeld2019generalized}
				& -- &55.8 &75.0  &63.9
				& -- &51.6 &53.5  &52.4
				& -- &47.2 &35.7  &40.6
				&-- &-- &-- &-- 
				&-- &-- &-- &--
				\\
				CE-ZSL \textit{CVPR 2021}~\cite{han2021contrastive}
				&70.4 &63.1 &78.6 &70.0
				&77.5 &63.9 &66.8 &65.3
				&63.3 &48.8 &38.6 &43.1
				&\textbf{\color{blue}70.6} &\textbf{\color{blue}69.0} &78.7 &73.5
				&--&--&--&--
				\\
				FREE \textit{ICCV 2021}~\cite{chen2021free}
				& -- &60.4 &75.4 &67.1
				& -- &55.7 &59.9 & 57.7
				& -- &47.4 &37.2 & 41.7
				& -- &67.4 &84.5 &75.0
				& -- & -- &-- & --
				\\
				HSVA \textit{NeurIPS 2021}~\cite{chen2021hsva}
				&-- &56.7 &79.8 &66.3
				&-- &52.7 &58.3 &55.3
				&-- &48.6 &39.0 &\textbf{\color{blue}43.3}
				&-- &-- &-- &-- 
				&-- &-- &-- &-- 
				\\
				\midrule
	
				CVC-ZSL \textit{ICCV2019}~\cite{li2019rethinking}
				&71.1 &56.4 &\textbf{\color{blue}{81.4}} &66.7
				&54.4 &47.4 &47.6 &47.5
				&62.6 &36.3 &\textbf{\color{red}42.8} &39.3
				&--&--&--&--
				&38.0 &26.5 &\textbf{\color{red}74.0} &39.0
				\\
				AREN \textit{CVPR 2019}~\cite{xie2019attentive}
				&67.9&54.7&79.1&64.7
				&71.8&38.9&{\color{blue}\textbf{78.7}}&52.1
				&60.6&19.0&38.8&25.5
				&--&--&--&--
				&-- &30.0 &47.9 &36.9
				\\
				DVBE \textit{CVPR 2020}~\cite{min2020domain}
				&--&62.7&77.5&69.4
				&--&64.4&73.2&68.5
				&--&44.1&\textbf{\color{blue}41.6} &42.8
				&--&--&--&--
				&-- &\textbf{\color{blue}37.9} &\textbf{\color{blue}55.9} &\textbf{\color{red}45.2}
				\\
				RGEN \textit{ECCV 2020}~\cite{xie2020region}
				&\textbf{\color{blue}73.6} &67.1 &76.5 &\textbf{\color{blue}71.5}
				&67.1 &60.0 &73.5 &66.1
				&63.8 &44.0 &31.7 &36.8
				&-- &-- &-- &--
				&\textbf{\color{blue}44.4} &\textbf{\color{red}48.1} &30.4 &37.2
				\\
				APN \textit{NeurIPS 2020}~\cite{xu2020attribute}
				&68.4 &56.5 &78.0 &65.5
				&72.0 &\textbf{\color{blue}65.3} &69.3 &67.2
				&61.6 &41.9 &34.0 &37.6
				&-- &-- &-- &--
				&-- &-- &-- &--
				\\
				GEM-ZSL \textit{CVPR 2021}~\cite{liu2021goal}
				&67.3&64.8&77.5&70.6
				&\textbf{\color{blue}77.8}&64.8& 77.1 &\textbf{\color{blue}70.4}
				&62.8&38.1&35.7&36.9
				&--&--&--&--
				&--&--&--&--
				\\
				
				\cline{1-21}
				\midrule
				\textbf{S2V$^*$ }
				&69.3  &61.0 &71.5 &65.8 
				&50.5  &41.8 &47.9 &44.7 
				&59.5  &43.1 &36.4 &39.5
				&47.3  &42.8 &67.7 &52.4
				&37.1  &31.2 &43.1 &36.2
				\\
				\textbf{S2V + EP \& EI }
				&72.7 &64.2 &76.5 &69.8
				&61.7 &54.1 &54.2 &54.1
				&63.5 &51.4 &35.9 &42.2
				&71.1 &63.8 &81.2 &71.5
				&43.5  &33.4 &57.2    &42.2
				\\
				\textbf{S2V + SoF}
				&72.7	&67.0	&79.6	&72.8
				&78.0	&71.9	&73.4	&72.6
				&64.3	&50.4	&36.8	&42.5
				&69.1	&62.5	&76.8	&68.9
				&40.2  &34.0 &52.5 &41.3
				\\
				\textbf{S2V + SoF \& $\textbf{EP}$}
				&77.3	&70.4	&80.1	&74.9
				&80.1	&76.0	&74.2	&75.1
				&65.9	&53.8	&35.6	&42.8
				&75.3	&69.5	&84.8	&76.3
				&42.4   &35.3   &53.5   &42.5
				\\
				\textbf{S2V + SoF \& $\textbf{EP}$ \& $\textbf{EI}$}
				&\textbf{\color{red}80.4}&\textbf{\color{red}74.6}&\textbf{\color{red} 82.6}&\textbf{\color{red}78.4}
				&\textbf{\color{red}83.5}&\textbf{\color{red}76.0}&\textbf{\color{red}79.2}&\textbf{\color{red}77.6}
				&\textbf{\color{red}68.7}&\textbf{\color{red}54.9}&35.9&\textbf{\color{red}43.4}
				&\textbf{\color{red}77.9}&\textbf{\color{red}73.7}&\textbf{\color{blue}86.8}&\textbf{\color{red}79.7}
				&\textbf{\color{red}44.9} &36.7 &53.4 &\textbf{\color{blue}43.5}
				\\
				\bottomrule[2pt]
				
			\end{tabular}
			
		}
		\caption{Comparison on five datasets. The first part is generative methods, the second part is feature generation methods. The best and second-best results are marked in {\color{red}\textbf{red}} and {\color{blue}\textbf{blue}}, respectively.  
		$^*$  denotes the reproduced results built on the source code of CVC-ZSL. EP, EI and SoF denote embedding propagation, embedding interpolation and semantic-oriented fine-tuning, respectively.
		} 
		\label{Table1}
	\end{table*}

\subsection{Datasets and Evaluation Protocol}

We conduct extensive experiments on five benchmark datasets: AWA2
~\cite{xian2017zero},
CUB
~\cite{welinder2010caltech}, 
SUN
~\cite{patterson2012sun}, 
FLO
~\cite{nilsback2008automated},
and APY
~\cite{farhadi2009describing}.
For class semantic embeddings, we use manually-labeled attribute vectors for AWA2, CUB, SUN, and APY, and use 1024-dimensional semantic embeddings extracted from text descriptions~\cite{reed2016learning} for FLO dataset.

We follow the datasets split and evaluation protocol proposed by ~\cite{xian2017zero}.
Under the conventional ZSL setting, we evaluate the per-class Top-1 accuracy on unseen classes, denoted as T. Under the GZSL setting, we evaluate the Top-1 accuracy on seen classes and unseen classes, respectively denoted as S and U. The holistic performance of GZSL is measured by their harmonic mean: $H = (2 \times U \times S)/(U + S)$.


\subsection{Implementation Details} \label{Implementation Details}

We adopt ResNet-101 pre-trained on ImageNet as the backbone.  
For training of SoF, we adopt the SGD optimizer with batch size of 16. During
training of semantic mapping network, we use Adam optimizer.
We also adopt an episode-based training fashion to sample $M$ categories and $N$ images for each category in a mini-batch. 
For all datasets, we obtained hyper-parameters by grid search on the validation set \cite{xian2017zero}. We set $M = 20$ and $N = 4$ and set hyperparameters $\sigma = 0.2$, ($\alpha_1, \alpha_2) = (5, 1)$. 
During the inference of GZSL, we use the calibrated stacking
for balanced performance.
Refer to Appendix A in supplementary file for more details.

\renewcommand\arraystretch{1.2}
\begin{table}[ht]
		\centering
		\resizebox{0.48\textwidth}{!}
		{
			\begin{tabular}{l| l | cccc |cccc }
				\toprule[2pt]
				\multirow{2}*{Methods}
				&\multirow{2}*{FT}
				&\multicolumn{4}{c|}{AWA2}
				&\multicolumn{4}{c}{CUB}     
		         \\
				&  & $T$ & ${U}$ & ${S}$ & ${H}$  & $T$ & ${U}$ & ${S}$ & ${H}$
				 \\
				 
				\midrule
				\multicolumn{1}{l|}{\multirow{2}{*}{TCN \cite{jiang2019transferable} +}} &  VF 
				&68.6 &61.2 &67.9 &64.4
				&72.5 &65.3 &69.0 &67.1 \\
				\multicolumn{1}{l|}{}  & SoF 
				&71.5 &67.2 &66.4 &66.8
				&77.6 &70.1 &73.6 &71.8
				\\
				\midrule 
				\multicolumn{1}{l|}{\multirow{2}{*}{CN\cite{skorokhodov2020class} +}} &  VF 
				&59.3 &53.2 &77.8 &63.2
				&65.4 &50.9 &79.1 &61.9 \\
				\multicolumn{1}{l|}{}  & SoF 
				&64.9 &61.6 &71.8 &66.3
				&76.1 &67.3 &78.9 &72.6
				\\
			        
			    \midrule
				\multicolumn{1}{l|}{\multirow{2}{*}{TF-VAEGAN\cite{narayan2020latent} +}} &  VF 
				&74.6 &58.1 &\textbf{\color{red}87.3} &69.8
				&76.7 &68.2 &75.0 &71.4 \\
				\multicolumn{1}{l|}{}  & SoF 
				&76.7 & 62.2 & 86.6 &72.4
				&80.2 & 71.2 & 79.2 &74.9 
				\\
			    	
				\midrule
				\multicolumn{1}{l|}{\multirow{2}{*}{LPL + }} &  VF 
				&76.1 & 67.9 &81.8 &74.2
				&76.8 & 70.6 &73.7 &72.1 \\
				\multicolumn{1}{l|}{}  &  SoF 
				&\textbf{\color{red}80.4} 
				&\textbf{\color{red}74.6}  
				&82.6
				&\textbf{\color{red}78.4}
				&\textbf{\color{red}83.5}
				&\textbf{\color{red}76.0}
				&\textbf{\color{red}79.2}
				&\textbf{\color{red}77.6}
				\\
				\bottomrule[2pt]
			\end{tabular}
		}
        \caption{Comparison between different fine-tuning paradigms. VF denotes vanilla fine-tuning with one-hot class labels and SoF denotes the proposed semantic-oriented fine-tuning. 
        \label{Table2}
		}
	\end{table}

\subsection{Comparison with SOTA}
To further demonstrate the superiority of our model, we compare LPL with recent methods under inductive settings as shown in Table \ref{Table1}.  We categorize the compared methods into generative and non-generative methods. Obviously, LPL achieves significant improvement. 
Especially on CUB dataset, LPL outperforms all the compared methods with a large margin, e.g., $7.4\%$, $6.0\%$ and $6.3\%$ for ZSL comparative to RGEN~\cite{xie2020region}, GEM-ZSL~\cite{liu2021goal}, CE-ZSL~\cite{han2021contrastive}. Considering CUB is a fine-grained dataset and requires discriminative prototypes for recognition, the impressive performance sufficiently demonstrates the effectiveness of learning prototypes via placeholders. 
For AWA2, LPL can also achieve an amazing result, which indicates that our model is applicable to datasets with different granularity. 
Besides, LPL does not achieve significant gains on SUN and APY datasets. 
SUN owns a large number of categories and too few samples in each category.
Also, the two datasets include much background information which restricts hallucinated placeholders from playing a effective role.


\begin{figure}[!t]
\centering
    \subfigure[{CVC-ZSL}]{ 
    \includegraphics[width=0.42\columnwidth]{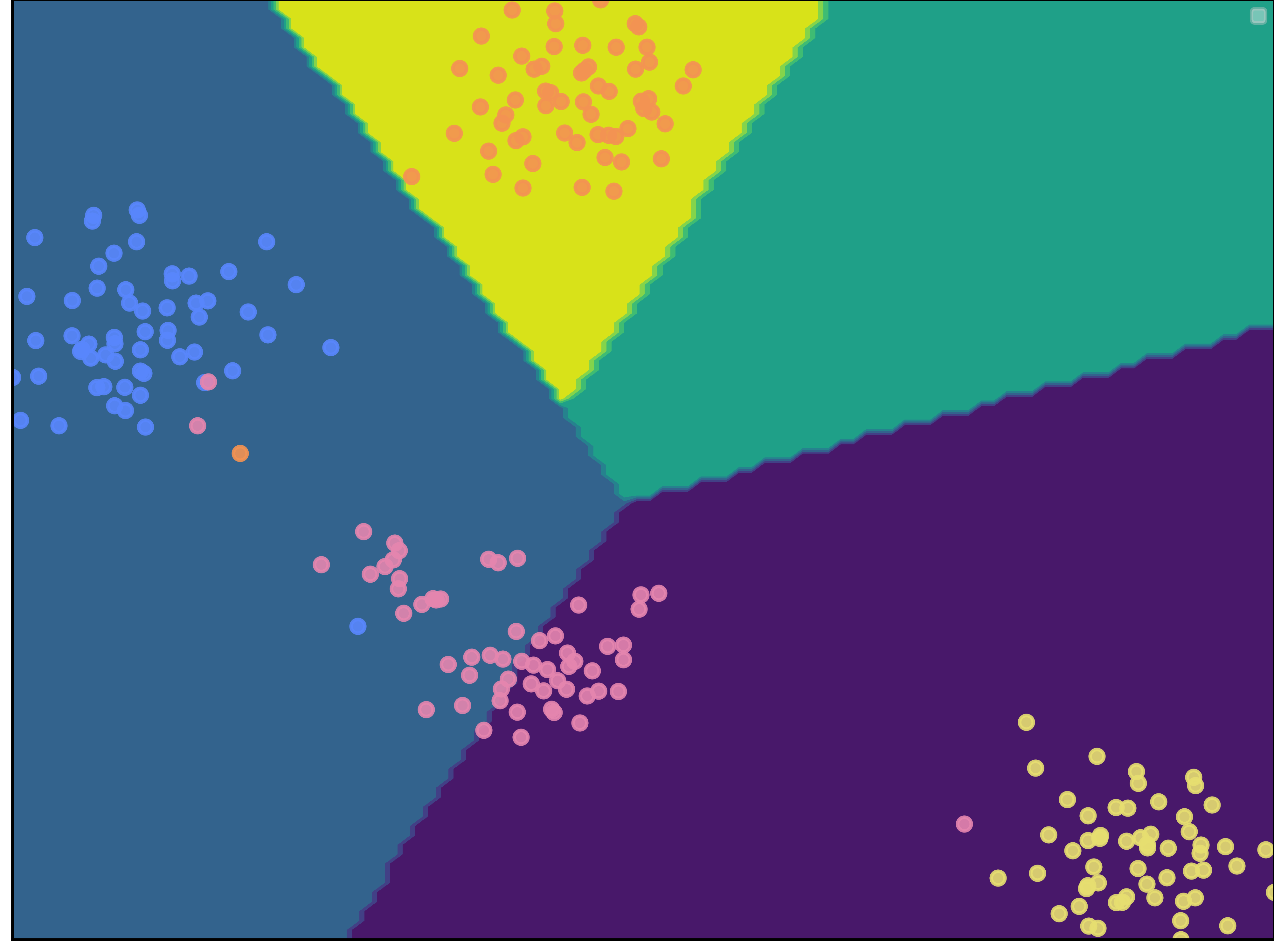} }
    \subfigure[{LPL}]{ 
    \includegraphics[width=0.42\columnwidth]{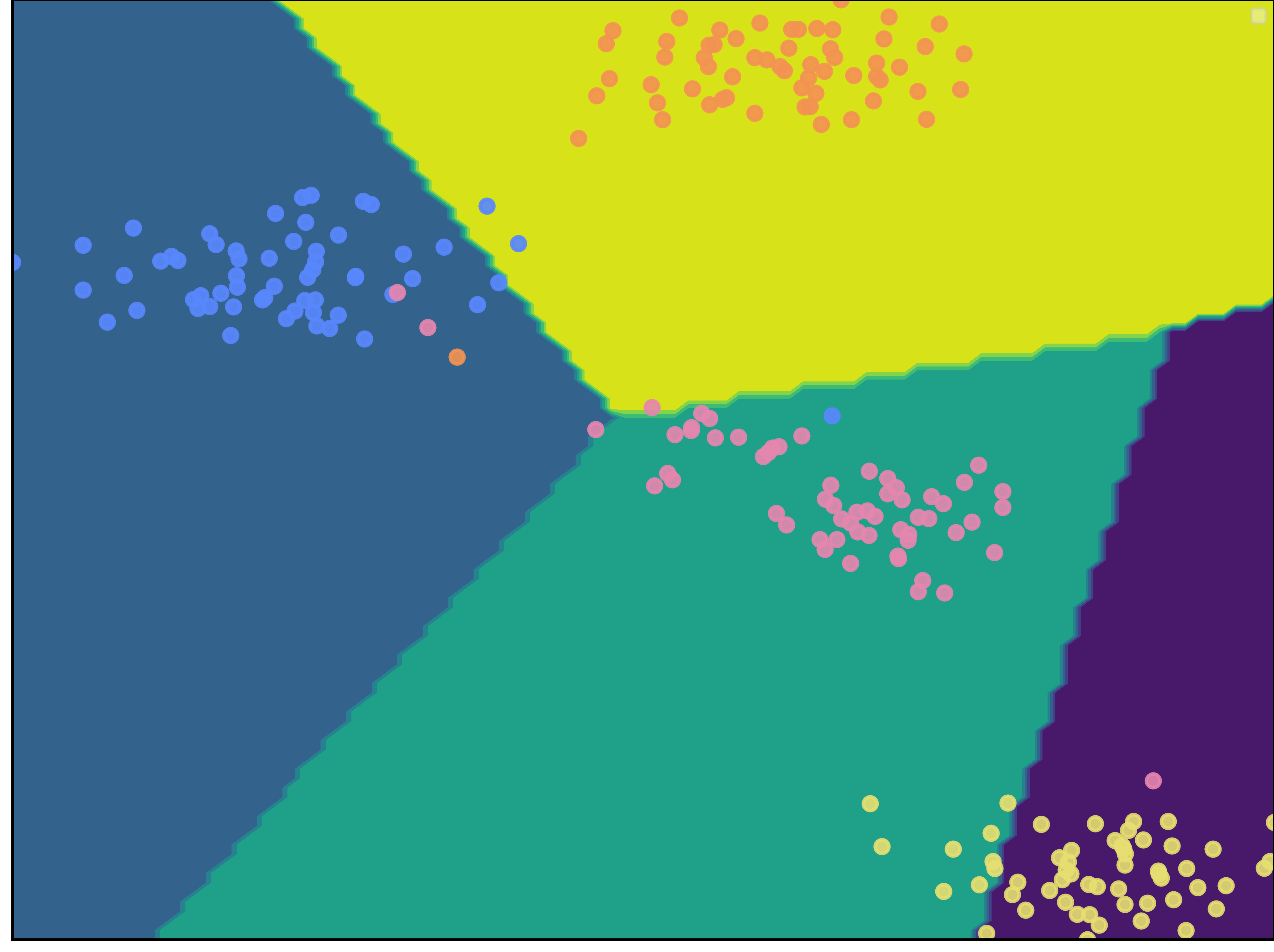} }
\caption{Visualization of decision boundaries on CUB unseen classes. With well-separated classes prototypes, LPL obtains more robust decision boundaries to counter the domain shift.}
\label{Figure4}
\end{figure}

\begin{figure*}[t]
\centering
    \subfigure[{Seen (CVC-ZSL)}]{ 
    \includegraphics[width=0.45\columnwidth]{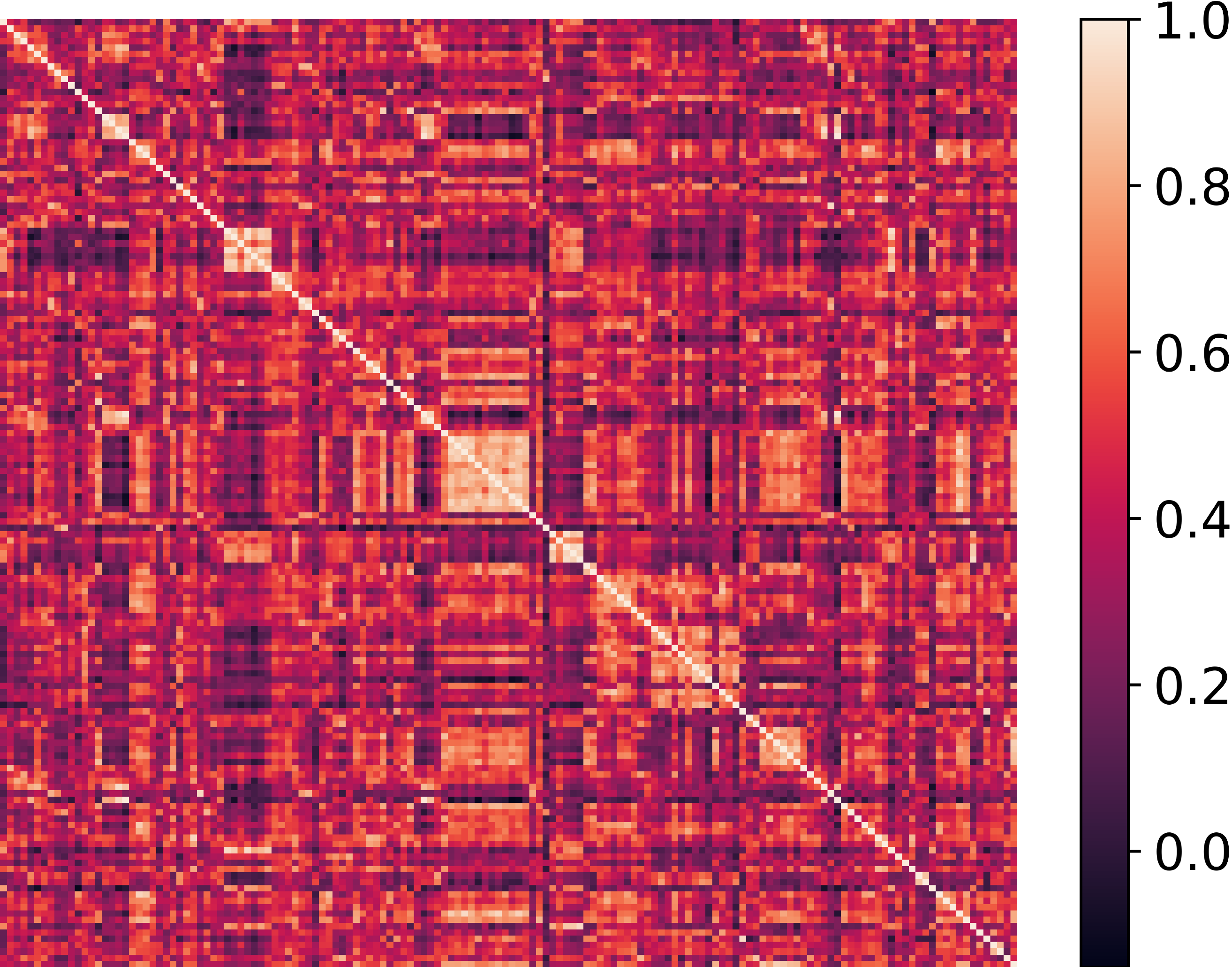} }
    \subfigure[{Seen (LPL)}]{ 
    \includegraphics[width=0.46\columnwidth]{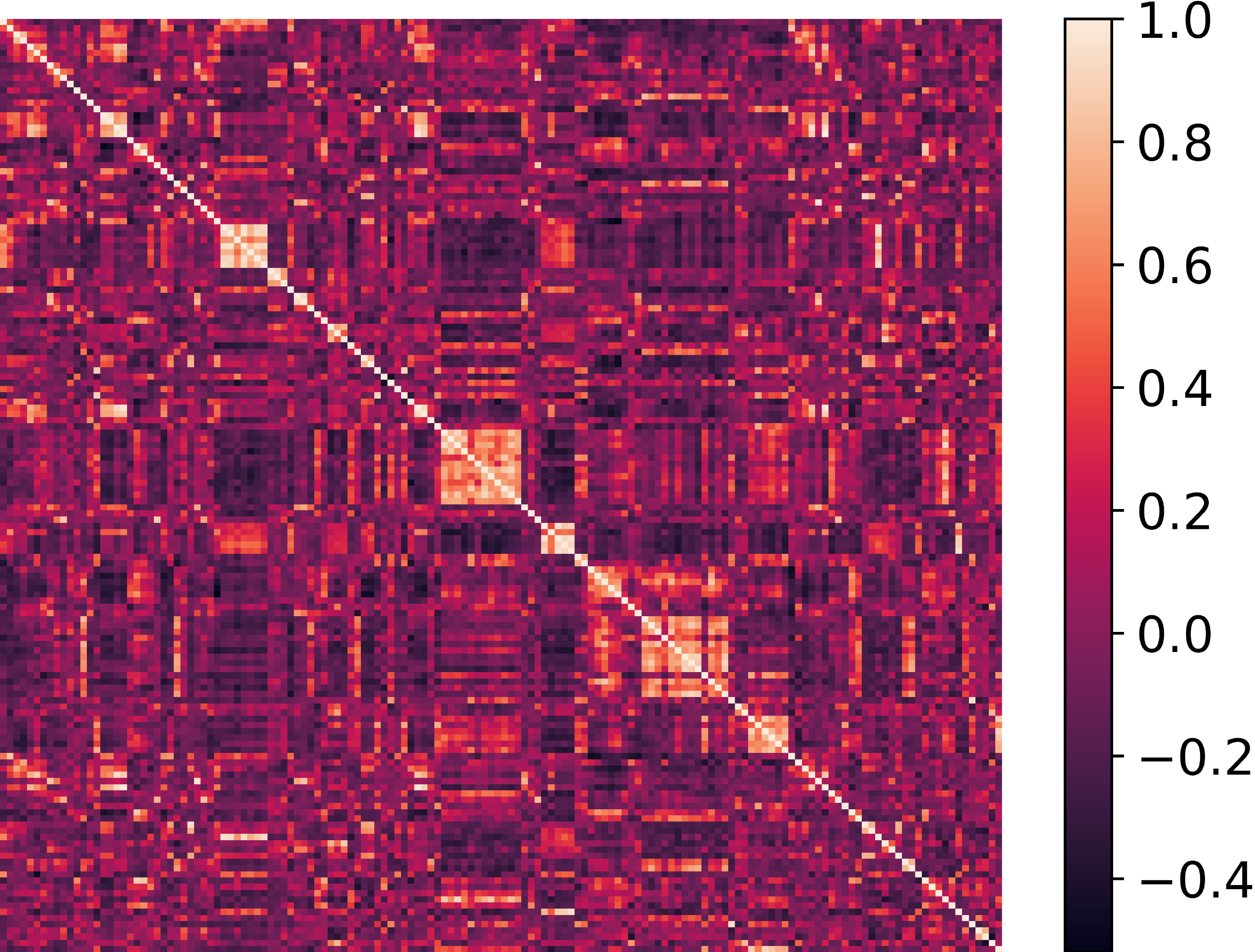} }
    \subfigure[{Unseen (CVC-ZSL)}]{ 
    \includegraphics[width=0.45\columnwidth]{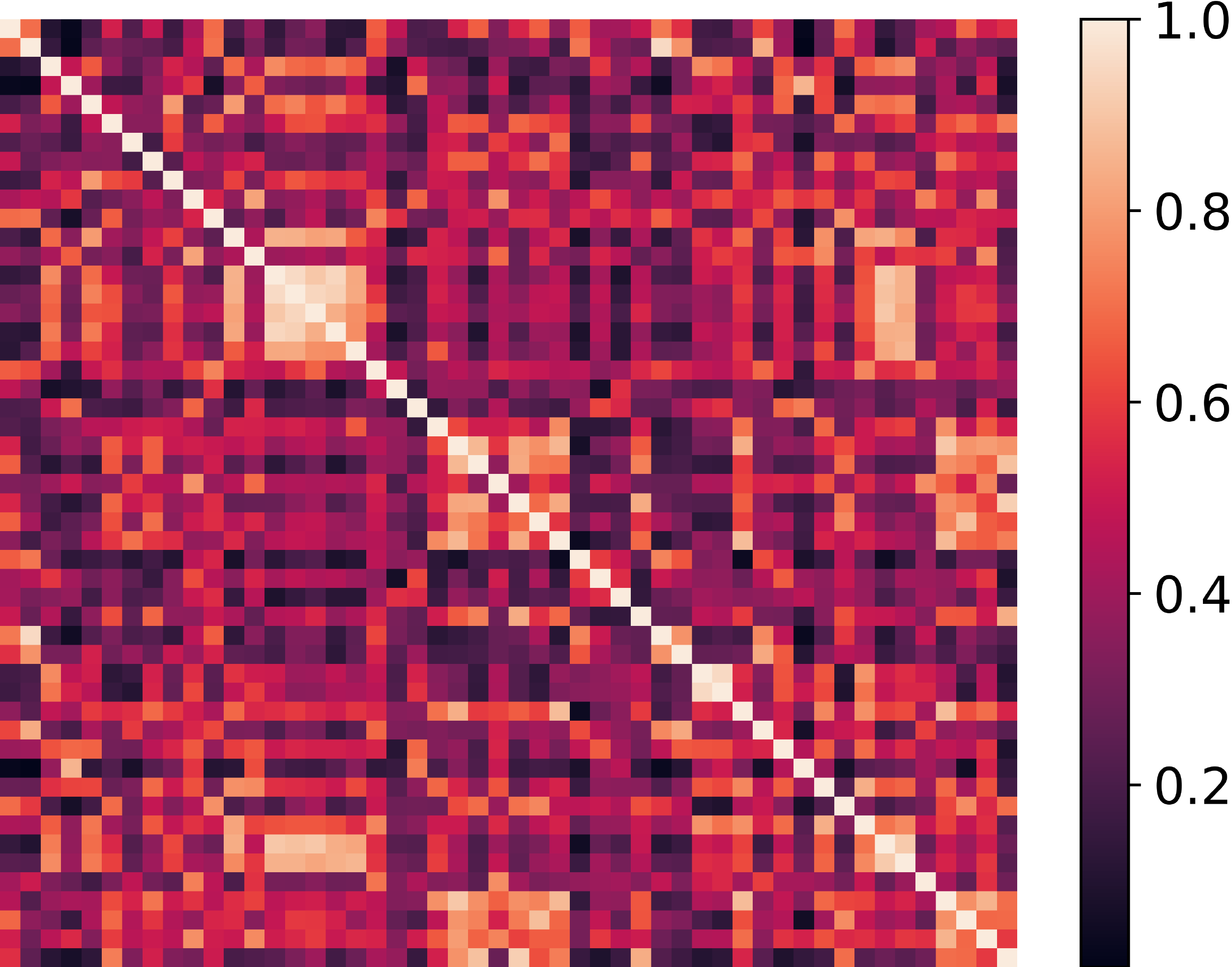} }
    \subfigure[{Unseen (LPL)}]{ 
    \includegraphics[width=0.46\columnwidth]{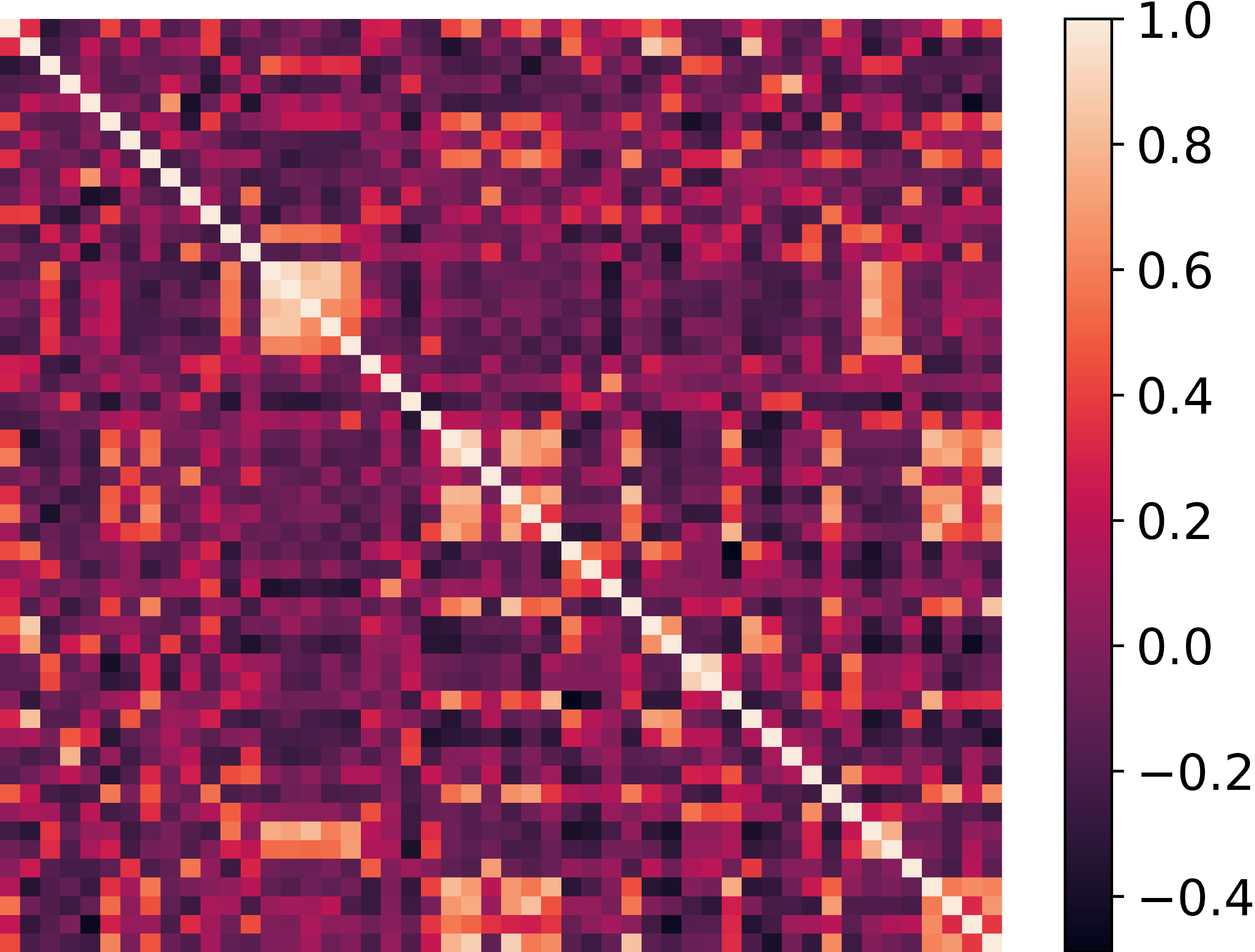} }
\caption{The heat-map of cosine similarity between prototypes learned by CVC-ZSL (baseline) and LPL on CUB dataset. The darker the color indicates that the prototypes are less similar and better separated in visual space. In particular, the similarity of the prototype to itself is $1.0$ on the diagonal line.}
\label{Figure3}
\end{figure*}

\subsection{Ablation Studies} \label{sec:ablation}

\noindent\textbf{Component Analysis.} 
The ablation results are shown at the bottom of Table \ref{Table1}. 
Firstly, we reproduce CVC-ZSL, i.e, a plain semantic$\rightarrow$visual (S2V) mapping network, as the baseline. To validate the effect of placeholders, we directly hallucinate classes built on raw visual embeddings. It could be seen that consistent improvement is achieved. Then, we fine-tune the backbone by semantic-oriented fine-tuning (SoF) for refined visual representations. S2V obtains obvious gains especially $4.0\%$ on AWA2 and $9.4\%$  on CUB. 
Next, built on refined visual representations, we introduce placeholders by two steps. First, we preliminarily hallucinate classes as placeholders by embedding propagation (EP). Then more hallucinated classes are produced by embedding interpolation (EI). We observe obvious improvement on five ZSL benchmarks respectively.
Overall, introducing placeholders ~(EP \& EI) contributes to learning discriminating prototypes and boosts the ZSL accuracy by $7.7\%$~(AWA2), $5.5\%$~(CUB), and $8.8\%$~(FLO).
The ablation experimental results fully validate the effectiveness of our ideas.


\noindent\textbf{Discussion about SoF.} 
We fine-tune CNN backbone by semantic embeddings to extract semantic-related visual representations. 
Previous work~\cite{xian2019f,narayan2020latent} also fine-tune backbone by class labels, named vanilla fine-tuning (VF) here. For fair comparison, in Table \ref{Table2}, we evaluate performance of TF-VAEGAN~\cite{narayan2020latent}, TCN~\cite{jiang2019transferable}, CN~\cite{skorokhodov2020class} with refined visual representations by SoF. Despite showing consistent improvement, LPL still achieves SOTA performance and demonstrates the superiority of semantic-oriented fine-tuning and learning prototypes via placeholders. 
Besides, when LPL hallucinates with visual representations refined by VF, it shows a slightly decreased performance, which demonstrates the necessity of prior visual-semantic alignment instead of a plain clustering.

\begin{figure}[!t]
\centering
    \subfigure[{Raw/CUB}]{ 
    \includegraphics[width=0.46\columnwidth]{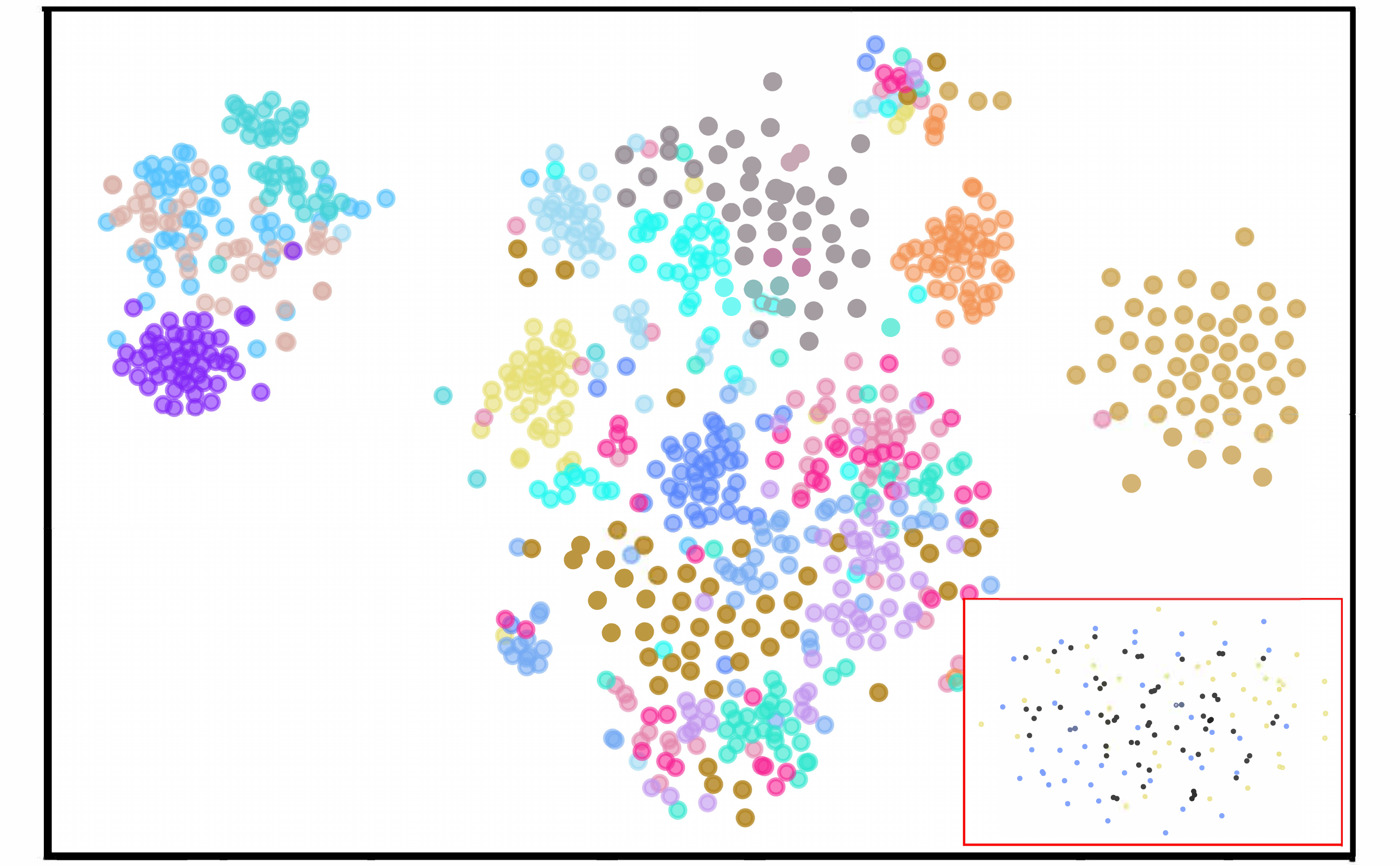} }
    \subfigure[{SoF/CUB}]{ 
    \includegraphics[width=0.46\columnwidth]{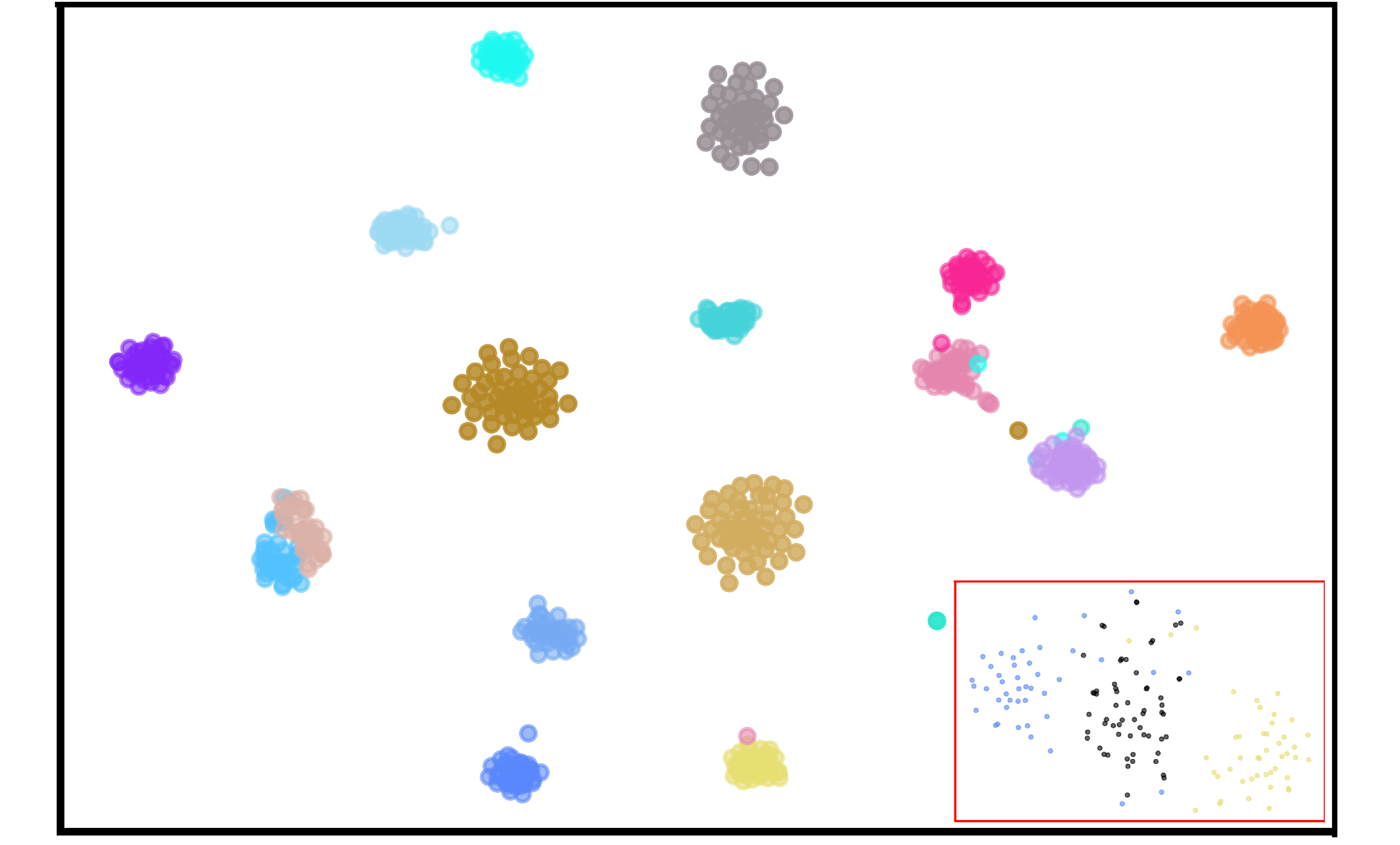} }
\caption{The t-SNE visualization of raw features (a) and refined features (b). For the latter, less semantic ambiguity is produced for hallucinated classes (with black color).}
\label{TSNE}
\end{figure}

\subsection{Visualization Analysis} \label{sec:visualization}

\noindent\textbf{Visualization of Well-separated Prototypes.} \label{Well-separated Prototypes}
To showcase the effectiveness of LPL in learning well-separated class prototypes, we calculate cosine similarity between class prototypes and  visualize the similarity heat-maps.
As shown in Figure \ref{Figure3}, (a)-(b) display the similarity between all seen classes prototypes on CUB and (c)-(d) display the similarity between all unseen ones.
We could find that the similarity, not only seen but also unseen, is significantly reduced, which means that the prototypes have a more discrete distribution as our declared motivation. 
The results shown in Figure \ref{Figure4} agree with the quantitative results in Table \ref{Table1}. 
When more space is spared for the placement of unseen classes prototypes, the decision boundaries built on unseen domain tend to be less ambiguous when facing semantically indistinguishable categories and lead to better classification accuracy. 

\noindent\textbf{Visualization of Refined Features by SoF.} 
We visualize the features refined by SoF on CUB dataset in Figure \ref{TSNE}. 
After fine-tuned by SoF, visual features obviously show better intra-class compactness with removing many semantic-unrelated features.
Another important point is that semantic ambiguity is prevented as much as possible during hallucinations. As shown in the bottom right of Figure \ref{TSNE} (a)-(b), hallucination classes data (with black color) have much overlap with genuine classes, which leads to semantic ambiguity, while the overlap is greatly avoided when visual embeddings are refined by SoF. 
More analysis is presented in Appendix B.



\subsection{Hyper-parameters Analysis} \label{sec:Hyper-parameters}

We evaluate the effect of different numbers of classes for hallucinations. 
We set number $n$ in $\{0, 1, 2, 3, 4, 5, 6, 7, 8\}$. The model is generally robust to the value of $n$.  When $n$ is too small, 
the performance is limited by the diversity of placeholders. When too large, the performance shows a slight decrease considering too many classes would increase the extent of semantic unreliability compared to authentic categories.
In order to balance diversity and faithfulness, we set $n=4$ for our all experiments.

We also evaluate the influence of the scaling factor $\sigma$. We consider $\sigma$ in $\{0.1, 0.2, 0.5, 1.0, 5.0, 10.0\}$. 
With the different $\sigma$ values, the Top-1 accuracy of ZSL and harmonic mean results on CUB datasets change slightly, indicating that our method is robust to the scaling factor. 
Specially, for all datasets in our work, we set $\sigma = 0.2$.
Corresponding figures for display and  more hyper-parameters analysis are presented in Appendix C.

\section{Conclusion}
In this paper, we propose an effective ZSL model termed LPL. With hallucination classes playing as placeholders of the unseen domain, the learned prototypes are encouraged to be well-separated. The hallucination method can obtain abundant classes with less cost by embedding propagation and interpolation on graphs composed of visual and semantic embeddings of seen classes.
Furthermore, semantic-oriented fine-tuning is adopted to decrease substandard hallucination data. 
Small to medium-sized datasets are considered in our work, larger data scenarios will be explored in the future.

\section*{Acknowledgments}
This study is partially supported by National Natural Science Foundation of China (62176016), the National Key R\&D Program of China (No.2021YFB2104800, No. 2018YFB2101100 and No.2019YFB2101600).
Guizhou Province Science and Technology Project: Research and Demonstration of Sci.\& Tech Big Data Mining Technology Based on Knowledge Graph (supported by Qiankehe[2021] General 382).


\bibliographystyle{named}
\bibliography{egpaper_final}

\end{document}